\documentclass[11pt,a4paper]{article}
\usepackage[hyperref]{acl2020}
\usepackage{times}
\usepackage{latexsym}

\usepackage{url}

\usepackage{amsmath,amsfonts,bm}

\def\eqref#1{equation~\ref{#1}}
\def\1{\bm{1}}

\DeclareMathAlphabet{\mathsfit}{\encodingdefault}{\sfdefault}{m}{sl}
\SetMathAlphabet{\mathsfit}{bold}{\encodingdefault}{\sfdefault}{bx}{n}

\usepackage{url}

\usepackage{amssymb}
\usepackage{wrapfig,lipsum,booktabs}
\usepackage{amsmath}
\usepackage{cleveref}
\usepackage{xcolor}
\usepackage{booktabs}
\usepackage{makecell}
\usepackage{bm}
\usepackage{algorithm}
\usepackage{algpseudocode}
\usepackage{graphicx}
\usepackage{nccmath}
\usepackage{longtable}
\usepackage{caption}
\usepackage{makecell}
\usepackage{subcaption}
\usepackage{tabularx}
\newcommand{\B}[1] {\boldsymbol{#1}}

\def\bW{{\B{W}}}

\def\bb{{\B{b}}}

\def\bc{{\B{c}}}

\def\bh{{\B{h}}}
\def\bm{{\B{m}}}

\def\bp{{\B{p}}}

\def\br{{\B{r}}}

\def\by{{\B{y}}}

\title{Boosting Naturalness of Language in Task-oriented Dialogues via Adversarial Training}

\author{Chenguang Zhu \\
Microsoft Speech and Dialogue Research Group, Redmond, WA, USA \\
  \texttt{chezhu@microsoft.com} \\}

\aclfinalcopy

\begin{document}
\maketitle

\begin{abstract}
The natural language generation (NLG) module in a task-oriented dialogue system produces user-facing utterances conveying required information. Thus, it is critical for the generated response to be natural and fluent. We propose to integrate adversarial training to produce more human-like responses. The model uses Straight-Through Gumbel-Softmax estimator for gradient computation. We also propose a two-stage training scheme to boost performance. Empirical results show that the adversarial training can effectively improve the quality of language generation in both automatic and human evaluations. For example, in the RNN-LG Restaurant dataset, our model AdvNLG outperforms the previous state-of-the-art result by 3.6\% in BLEU.

\end{abstract}

\section{Introduction}
In task-oriented dialogues, the computer system communicates with the user in the form of a conversation and accomplishes various tasks such as hotel booking, flight reservation and retailing. In this process, the system needs to accurately convert the desired information, a.k.a. \textit{meaning representation}, to a natural utterance and convey it to the users (Table~\ref{table:first_example}). The quality of response directly impacts the user's impression of the system. Thus, there are numerous previous studies in the area of natural language generation (NLG) for task-oriented dialogues, ranging from template-based models \citep{rule,halogen} to corpus-based methods \citep{tgen,ralstm,sclstm,zhu_nlg}.

However, one issue yet to be solved is that the system responses often lack the fluency and naturalness of human dialogs. In many cases, the system responses are not natural, violating inherent human language usage patterns. For instance, in the last row of Table~\ref{table:first_example}, two pieces of location information for the same entity \textit{restaurant} should not be stated in two separate sentences. In another example in Table~\ref{table:example}, the positive review \textit{child friendly} and the negative review \textit{low rating} should not appear in the same sentence connected by the conjunction \textit{and}. These nuances in language usage do impact user's impression of the dialogue system, making the system response rigid and less natural.

To solve this problem, several methods use reinforcement learning (RL) to boost the naturalness of generated responses \citep{rl1,li_rl}. However, the Monte-Carlo sampling process in RL is known to have high variance which can make the training process unstable. \citet{li2015diversity} proposes to use maximum mutual information (MMI) to boost the diversity of language, but this criterion makes exact decoding intractable.

\definecolor{darkpastelgreen}{rgb}{0.01, 0.75, 0.24}
\definecolor{seagreen}{rgb}{0.18, 0.55, 0.34}
\definecolor{mountainmeadow}{rgb}{0.19, 0.73, 0.56}
\definecolor{forestgreen}{rgb}{0.13, 0.55, 0.13}
\definecolor{fireenginered}{rgb}{0.81, 0.09, 0.13}

\begin{table}[t]
\centering
\small
\begin{tabular}{l|l}
\toprule
Input & \makecell{name[Wildwood], eatType[restaurant], \\food[Indian], area[riverside],\\ familyFriendly[no], near[Raja Indian Cuisine]} \\
\midrule
\makecell{with\\adv.} & \makecell{Wildwood is an Indian restaurant \textbf{\textcolor{forestgreen}{in the}} \\ \textbf{\textcolor{forestgreen}{riverside area near Raja Indian Cuisine}}. \\ It is not family friendly.}\\
\midrule
\makecell{w/o\\adv.} & \makecell{Wildwood is a restaurant providing Indian food. \\It is \textbf{\textcolor{fireenginered}{located in the riverside.}}\\ It is \textbf{\textcolor{fireenginered}{near Raja Indian Cuisine}}.}\\
\bottomrule
\end{tabular}
\caption{Example of generated utterances from meaning representation input. Our model learns to put two pieces of location information in one sentence via adversarial training.} 
\label{table:first_example}
\end{table}

On the other hand, the adversarial training for natural language generation has shown to be promising as the system needs to produce responses indiscernible from human utterances \citep{adv_nlg,advmt,relgan}. Apart from the generator, there is a discriminator network which aims to classify system responses from human results. The generator is trained to fool the discriminator, resulting in a min-max game between the two components which boosts the quality of generated utterances \citep{gan}. %As shown in Table~\ref{table:first_example}, with adversarial training, our model learns to group two pieces of location information together, which is more natural.
Due to the discreteness of language, most previous work on adversarial training in NLG apply reinforcement learning, suffering from high-variance problem \citep{seqgan, li2017adversarial,araml}.

In this work, we apply adversarial training to utterance generation in task-oriented dialogues and propose the model AdvNLG. Instead of using RL, we follow \citet{style} to leverage the Straight-Through Gumbel-Softmax estimator \citep{gumbelsoftmax} for gradient computation. In the forward pass, the generator uses the $\mbox{argmax}$ operation on vocabulary distribution to select an utterance and sends it to the discriminator. But during backpropagation, the Gumbel-Softmax distribution is used to let gradients flow back to the generator. %Unlike reinforcement learning, there is no sampling in the computation, which greatly minimizes the uncertainty in training. 
We also find that pretraining the generator for a warm start is very helpful for improving the performance.

To evaluate our model, we conduct experiments on public datasets E2ENLG \citep{e2enlg} and RNN-LG \citep{multidomain}. Our model achieves strong performance and obtains new state-of-the-art results on four datasets. For example, in Restaurant dataset, it improves the best result by 3.6\% in BLEU. Human evaluation corroborates the effectiveness of our model, showing that the adversarial training against human responses can make the generated language more accurate and natural.

\section{Problem Formulation}
The goal of natural language generation module in task-oriented dialogues is to produce system utterances directly issued to the end users \citep{young2000}. The generated utterances need to carry necessary information determined by upstream dialogue modules, including the \textit{dialogue act} (DA) and \textit{meaning representation} (MR).

%Examples include \textit{inform}, \textit{request} and \textit{confirm}. Different dialogue acts determine different forms of language. For instance, the \textit{request} act is usually associated with a question from the system, while the \textit{inform} act often corresponds to an informative statement.

The dialogue act specifies the type of system response (e.g. \textit{inform}, \textit{request} and \textit{confirm}), while the meaning representation contains rich information that the system needs to convey to or request from the user in the form of slot-value pairs. Each slot indicates the information category and each value represents the information content.% For example, (slot:\textit{food}, value:\textit{Italian}) means that the system needs to generate utterance indicating the restaurant serves Italian food. %When the dialogue act is \textit{request}, the meaning representation may consist of only slots, and the system needs to solicit the value from the user. We show some examples of DA, MR and utterances in Table~\ref{table:desp}.

Therefore, the training data for the supervised NLG task is $\{x_i=(d_i, r_i), y_i\}_{i=1}^n$, where $d_i$ is the dialogue act, $r_i=\{(s_1, v_1), ..., (s_t, v_t)\}$ is the set of MR slot-value pairs, and $y_i$ is the human-labeled response. %During inference, the system needs to generate a natural response given dialogue act $d$ and meaning representation $r$.

NLG models typically use \textit{delexicalization} during training and inference, replacing slots and values in the utterance with a special token $\langle$\textit{SLOT NAME}$\rangle$. In this way, the system does not need to generate the proper nouns. Finally, the model substitutes these special tokens with corresponding values when delivering to users. 

% \begin{table}[t]
% \centering
% \begin{tabular}{lc}
% \toprule
%  DA: inform\\
%  MR: \{(Food, Italian), (Area, City south)\} \\
%  \midrule
%  Utt.: There is a restaurant serving Italian food\\ 
%  in the South of the city.\\
%  \midrule
%  DA: request, MR: \{(Area)\} \\
%  \midrule
%  Utt.: Can I know where do you want to have\\
%   dinner?\\
% \bottomrule
% \end{tabular}
% \caption{Example dialogue act (DA), meaning representation (MR) and corresponding system utterances.} 
% \label{table:desp}
% \end{table}

\section{Model}
\subsection{Generator Model}
We use the sequence-to-sequence encoder-decoder architecture \citep{seq2seq} for the response generator $G$. The input to the encoder is a single sequence $x$ of length $m$ via concatenating dialogue act $d$ and slots and values in the meaning representation $r$. The target utterance $y$ has $n$ tokens, $y_1,...,y_n$. Following \citet{zhu_nlg}, we delexicalize both sequences and surround each sequence with $\langle$BOS$\rangle$ and $\langle$EOS$\rangle$ tokens.

\begin{figure*}[htbp]
\centering
\vspace{2.5\baselineskip}
\includegraphics[scale=0.52,trim=6cm 6cm 6cm 4cm]{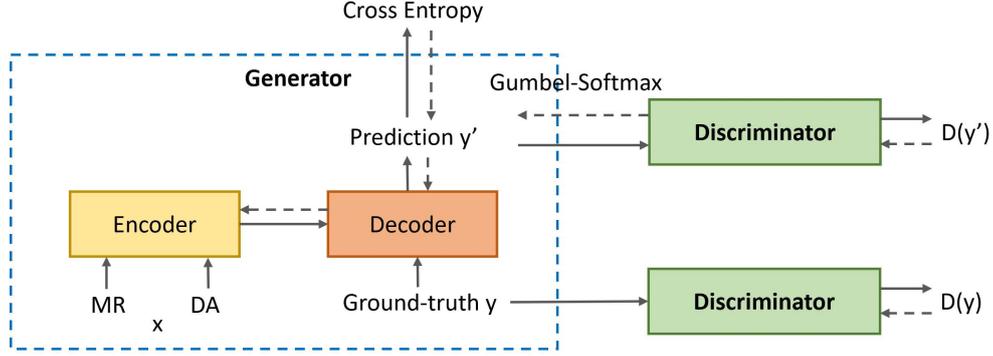}
\vspace{-5.5\baselineskip}
\caption{AdvNLG model with generator and discriminator. The dashed arrow is the direction of gradient flow.}
\label{fig:model}
\end{figure*}

%The training objective is to maximize the following conditional probability $p(y_1, ..., y_n|x) = \prod_{t=1}^n p(y_t|y_1,...,y_{t-1};x)$.

Both the encoder and decoder use GRU \citep{gru} for contextual embedder, and they share the embedding matrix $\mathbf{E}$ to map each token to a fixed-length vector. The final hidden state of the encoder RNN is used as the initial state of the decoder RNN. Moreover, the decoder employs a dot-product attention mechanism \citep{attention} over the encoder states to get a context vector $\bc$ at each decoding step. 

%Specifically, suppose the RNN's output states of the encoder are $(\bu_1, \bu_2, ..., \bu_m)\in \mathbb{R}^{d_h\times m}$, where $d_h$ is the dimension RNN's hidden state. Then, in the $t$-th step of the decoder, the dictionary $\mathcal{D}$ maps the $t$-th output token into $\bs_t$ and apply dropout. The hidden state of the previous RNN unit is $\bh_{t-1}$. The attention weights $\{\alpha_i\}_{i=1}^m$ are computed as follows:

%\begin{align}
%    \bv_i &= [\bh_{t-1}; \bu_i] \in \mathbb{R}^{2d_h} \\
%    \be_i &= \softmax(\bW_1\bv_i) \in \mathbb{R}^{d_h} \\
%    \alpha_i &= \relu(\bb^T \be_i)\in \mathbb{R},
%\end{align}
%where $\bW_1\in \mathbb{R}^{d_h \times 2d_h}$ and $\bb\in \mathbb{R}^{d_h}$ are learnable parameters.  The attention weights are then employed to obtain the context vector $\bc=\sum_{1\leq i \leq m}\alpha_i \bu_i$.

This context vector $\bc$ is concatenated with the embedding of the current token and fed into the GRU to predict the next token. 
% \begin{align}
%     \bo &= \bW_2[\bg; \bc]\in\mathbb{R}^d \\
%     \bp_t &= \softmax(\bW_\mathcal{D}\bo) \in \mathbb{R}^{|V|},
% \end{align}
% where $\bW_2\in \mathbb{R}^{d \times 2d_h}$ is a parametrized matrix, and $\bW_\mathcal{D}$ is the transposed weights of dictionary $\mathcal{D}$. 
The result $\bp_t=p(y_t|y_1,...,y_{t-1};x)$ is the probability distribution of the next token over all tokens in dictionary $V$. 

We use cross entropy as the generator's loss function. Suppose the one-hot ground-truth token vector at the $t$-th step is $\by_t$, then the loss is:

\begin{equation}
\label{eq:lgen}
    \mathcal{L}_{\mbox{Gen}}(\theta) = -\sum_{t=1}^n \by_t^T \log (\bp_t)
\end{equation}

\subsection{Adversarial Training}
The goal of the adversarial training is to use a discriminator to differentiate between the utterance $y'$ from generator and the ground-truth utterance $y$. 

We leverage the improved version of generative adversarial network (GAN), Wasserstein-GAN (WGAN) \cite{wgan}, in our framework. WGAN designs a min-max game between the generator $G$ and the discriminator $D$:

\begin{equation}
\label{eq:minmax}
    \min_{G}\max_{D} E_{y\sim P_{data}(y)}[D(y)] - E_{y'\sim G(x)}[D(y'))] 
\end{equation}
%\begin{align*}
%    \min_{G}\max_{D} & E_{y\sim P_{data}(y)}[\log D(y)] + \\
%     & E_{y'\sim G(x)}[\log (1 - D(y'))],
%\end{align*}
where $G(x)$ denotes the probability distribution computed by the generator $G$ given input $x$. The discriminator function $D$ is a scoring function on utterances.

The goal of the generator is to obtain $y'$ as similar as possible to $y$ to fool the discriminator $D$ (the outer-loop min), while $D$ learns to successfully classify generated output $y'$ from the ground-truth $y$ (the inner-loop max), via the scoring function $D$.

\subsubsection{Discriminator Model}
For the discriminator, we reuse the embedding matrix $\mathbf{E}$ as the embedder, followed by a bidirectional GRU layer. The last GRU hidden state $\bh$ is passed through a batch normalization layer and a linear layer to get the final score $D(y)$:

\begin{align}
    \br &= \mbox{BatchNorm}(\bh)\\
    D(y) &= \bW_{3}\br + \bb_3,
\end{align}
where $\bW_{3}$ and $\bb_{3}$ are trainable parameters.

\subsubsection{Training}
\textbf{Gradient computation.} One problem with adversarial training in language generation is that the token sequence $y'$ sampled from $G$ is discrete, making it impossible to back-propagate gradients from the min-max objective to the generator. %Most previous work on adversarial language generation \citep{advmt, adv_tod} employs reinforcement learning. But the involved sampling process introduces high variance that makes the training unstable. 

Several previous methods leverage reinforcement learning for gradient computation \citep{seqgan,li2017adversarial}. However, the related sampling process can introduce high variance during training. Therefore, we employ the Straight-Through Gumbel-Softmax estimator \citep{gumbelsoftmax,seq3}. In detail, during the forward pass, at the $t$-th step, the $\mbox{argmax}$ of the generated word distribution $\bp_t$ is taken, i.e. greedy sampling. But for gradient computation, the Gumbel-Softmax distribution is used as a differentiable alternative to the $\mbox{argmax}$ operation:

\begin{equation}
    \bp'_{t,i} = \frac{\exp(\log(\bp_{t,i}) + g_i)/\tau}{\sum_{j=1}^{|V|}\exp(\log(\bp_{t,j}) + g_j)/\tau},
\end{equation}
where $g_1, ..., g_{|V|}$ are i.i.d samples drawn from the Gumbel distribution $G(0, 1)$ and $\tau$ represents the softmax temperature. \citet{gumbelsoftmax} shows that the Gumbel-Softmax distribution converges to the one-hot distribution as $\tau \rightarrow 0$ and to the uniform distribution as $\tau \rightarrow \infty$. We set $\tau=0.1$ in all the experiments.

\textbf{Two-stage Training.} We find that the adversarial training does not work well if we optimize both the cross entropy (Eq.~\ref{eq:lgen}) and the min-max objective (Eq.~\ref{eq:minmax}) from the beginning. However, after we warm up the generator model with only cross entropy loss for several epochs, and then train with the discriminator under both the cross entropy and adversarial objective, the performance is consistently boosted. We argue that during early stages, the generator cannot produce meaningful output, making the discriminator easy to overfit. It's then hard for generator to learn to fool the adversary.

We summarize our model AdvNLG and gradient computation process in Fig.~\ref{fig:model}.

\section{Experiments}
We conduct empirical tests on a number of benchmarks for task-oriented dialogues over a variety of domains such as restaurant booking, hotel booking and retail. The datasets include the E2E-NLG task \citep{e2enlg} with 51.4K samples, and the TV, Laptop, Hotel and Restaurant datasets from RNN-LG \citep{multidomain}, with 14.1K, 26.5K, 8.7K and 8.5K samples respectively. We use BLEU-4 \citep{bleu} for the automatic metric, computed by the official evaluation scripts from E2E-NLG and RNN-LG. 

\subsection{Baselines}
The baseline systems include TGen \citep{tgen}, SC-LSTM \citep{sclstm}, RALSTM \citep{ralstm}, Slug \citep{slug}, S2S+aug \citep{cylin} and NLG-LM \citep{zhu_nlg}. We also implement adversarial training using reinforcement learning in the same way as \citet{li2017adversarial}, denoted by RL. The generator in RL is warmed up in the same way as AdvNLG.

\subsection{Training Details}
In all experiments, the learning rate is 1e-3, the batch size is 20 and the beam width in inference is 10. According to WGAN, the discriminator's parameters are clipped at 0.1. We use RMSprop \citep{rmsprop} as the optimizer. Teacher forcing is used for training the generator, which means that the decoder is exposed to the previous ground-truth token. In warm-up phase, we train the generator for 2 epochs. In E2E-NLG dataset, the generator is updated 5 times before the discriminator is updated once, which is typical in GAN training \citep{advmt}. The hyper-parameters above are chosen based on performance on the dev set. %We train the generator for 2 epochs using cross entropy, then co-train the generator and discriminator using cross entropy and the adversarial objective.
Other hyper-parameters like dropout rate, dictionary dimension and RNN hidden size are the same with Table 3 in \citet{zhu_nlg}. 

For baseline models, we implemented NLG-LM \citep{zhu_nlg} and reproduced its results. We obtain the prediction results of Slug \citep{slug} from its open-source website.

\begin{table}[tbp]
\centering
\begin{tabular}{p{1.5cm}>{\centering\arraybackslash}p{0.7cm}>{\centering\arraybackslash}p{0.7cm}>{\centering\arraybackslash}p{0.7cm}>{\centering\arraybackslash}p{0.7cm}>{\centering\arraybackslash}p{0.7cm}}
\toprule
 Model & \textbf{E} & \textbf{TV} & \textbf{L} & \textbf{H} & \textbf{R}\\
 \midrule
TGen & 0.659  & / & / & / & /\\ 
Slug & 0.662 & 0.529 & 0.524 & / & /  \\
SCLSTM & / & 0.527 & 0.512 & 0.848 & 0.752 \\
RALSTM & / & 0.541 & 0.525 & 0.898 & 0.779 \\
S2S+aug & 0.665 & / & / & / & /\\ 
NLG-LM & \textbf{0.684} & 0.617 & 0.586 & 0.939 & 0.795 \\
\midrule
AdvNLG & \textbf{0.683} & \textbf{0.625$^{*}$} & \textbf{0.624$^{*}$} & \textbf{0.945$^{*}$} & \textbf{0.831$^{*}$} \\
\hspace{0.05cm} RL & 0.674 & 0.605 & 0.606 & 0.932 & 0.796 \\
\hspace{0.05cm} -Adv. & 0.671 & 0.618 & 0.564 & 0.931 & 0.753 \\
\hspace{0.05cm} -2 stages & 0.662 & 0.621 & 0.557 & 0.932 & 0.782 \\
\bottomrule
\end{tabular}
\caption{BLEU scores on E2ENLG (\textbf{E}), TV, Laptop (\textbf{L}),  Hotel (\textbf{H}) and Restaurant (\textbf{R}) testset. *: means the result is statistically significant with p-value$<$0.05. \textbf{-Adv.} means we only train the generator, with cross entropy loss. \textbf{-2 stages} means that both the generator and discriminator are trained together from scratch.} \label{table:mainresult}
\end{table}

\subsection{Results}
As shown in Table~\ref{table:mainresult}, our model AdvNLG achieves new state-of-the-art results on TV, Laptop, Hotel and Restaurant datasets, improving previous best results by 0.8\%, 3.8\%, 0.6\% and 3.6\%. Statistical tests show that this advantage is statistically significant with p-values smaller than 0.05. Our model also obtains results on par with NLG-LM on E2ENLG. We show some prediction examples in Table~\ref{table:example}. Generally, with adversarial training, the generated output can group information from the same category together, while placing positive and negative aspects (e.g. family-friendly and expensive) in different sentences. 

\textbf{Ablation Study.} %We evaluate the effectiveness of the proposed adversarial training. 
The bottom section of Table~\ref{table:mainresult} shows that adversarial training can boost performance by 0.7\% to 7.8\%. Our proposed two-stage training is also very beneficial. If both generator and discriminator are trained from scratch, the result drops significantly. RL-based adversarial training achieves mixed results. On TV dataset, it even hurts the performance. We attribute this to the high variance and instability in training.

\begin{table}[hbp]
\centering
\begin{tabular}{l|c|c}
\toprule
 Model & Naturalness & Accuracy\\
 \midrule
Slug & 2.51 (0.48) & 2.89 (0.36)    \\
NLG-LM & 2.52 (0.46) & 2.84 (0.41) \\
AdvNLG & \textbf{2.84$^*$} (0.27) & \textbf{2.97$^*$} (0.17) \\
\hspace{0.05cm} -Adv. & 2.45 (0.53) & 2.63 (0.58) \\
\bottomrule
\end{tabular}
\caption{Average human evaluation ratings (1-3, 3 is best) for naturalness and accuracy of output generated by different models. Standard deviation is shown in parenthesis. $^*$: the p-value is smaller than 0.01.} \label{table:human}
\end{table}

\begin{table*}[t]
\centering
\begin{tabular}{p{1.5cm}|p{13.2cm}}
\hline
MR & name[Wildwood], eatType[restaurant], food[Indian], area[riverside], familyFriendly[no], near[Raja Indian Cuisine] \\
Ref. &  Located in the riverside area near the Raja Indian Cuisine, Wildwood offers Indian food and a restaurant. It is not family friendly. \\
AdvNLG &  Wildwood is an Indian restaurant \textcolor{forestgreen}{in the riverside area near Raja Indian Cuisine}. It is not family friendly. \\
\hspace{0.05cm} -Adv. & Wildwood is a restaurant providing Indian food. It is located \textcolor{fireenginered}{in the riverside}. It is \textcolor{fireenginered}{near Raja Indian Cuisine}. \\
NLG-LM & Wildwood is a restaurant providing Indian food. It is located \textcolor{fireenginered}{in the riverside}. It is \textcolor{fireenginered}{near Raja Indian Cuisine}. \\
\hline
Comment & Only AdvNLG places the two pieces of location information ``riverside'' and ``near Raja Indian Cuisine'' together, which is aligned with human language patterns. \\
\hline
\hline
MR & name[The Cricketers], eatType[restaurant], food[English], priceRange[high], customer rating[1 out of 5], area[city centre], familyFriendly[yes], near[Café Rouge] \\
Ref. & The Cricketers, an English restaurant located near Café Rouge in the city centre, offers food at high price range. Although it has a customer rating of 1 out of 5, it also is children friendly. \\
AdvNLG & The Cricketers is a child friendly English restaurant in the city centre near Café Rouge. It has a \textcolor{forestgreen}{high price range} \textbf{and} a \textcolor{forestgreen}{customer rating of 1 out of 5}. \\
\hspace{0.05cm} -Adv. & The Cricketers is a restaurant located in the city centre near Café Rouge. It is a high priced restaurant that serves English food. It is \textcolor{fireenginered}{rated 1 out of 5} \textbf{and} is \textcolor{fireenginered}{children friendly}. \\
NLG-LM & The Cricketers is a high priced English restaurant located in the city centre near Café Rouge. It has a \textcolor{fireenginered}{customer rating of 1 out of 5} \textbf{and} is \textcolor{fireenginered}{child friendly}. \\
\hline
Comment & AdvNLG model naturally put the negative aspects like ``high price'' and ``rating 1 out of 5'' together with conjunction ``and'', whereas both -Adv. and NLG-LM juxtapose negative aspect (low customer rating) and positive aspect (kid-friendly) in one sentence, which appears contradictory.\\
\hline
\hline
MR & name[The Plough], eatType[restaurant], food[Chinese], priceRange[cheap], area[riverside], familyFriendly[yes], near[Raja Indian Cuisine] \\
Ref. & The Plough is a cheap Chinese restaurant located riverside by Raja Indian Cuisine. It is a family friendly establishment. \\
AdvNLG & The Plough is a cheap \textcolor{forestgreen}{Chinese restaurant} \textcolor{forestgreen}{in the riverside area near Raja Indian Cuisine}. It is family friendly. \\
\hspace{0.05cm} -Adv. & The Plough is a cheap family friendly \textcolor{fireenginered}{restaurant that serves Chinese food}. It is located in the riverside area near Raja Indian Cuisine. \\
NLG-LM & The Plough is a restaurant providing Chinese food in the cheap price range. It is located \textcolor{fireenginered}{in the riverside}. It is \textcolor{fireenginered}{near Raja Indian Cuisine}. \\
\hline
Comment & AdvNLG places ``Chinese'' immediately before ``restaurant'', and this is in line with the human reference. And NLG-LM model has two less connected sentences at the end.\\
\hline
\end{tabular}
\caption{Example of predictions on E2E-NLG by reference, NLG-LM model and our model AdvNLG with and without adversarial training. As E2E-NLG only has \textit{inform} dialogue act, we show the meaning representation (MR).}
\label{table:example}
\end{table*}

\subsection{Human Evaluation}
We randomly sample 100 data-text pairs from the test set of E2ENLG. We then ask 3 labelers to judge the accuracy and naturalness of the utterances generated by Slug, NLG-LM, AdvNLG with and without adversarial training. The accuracy measures how precisely the utterance expresses the dialogue act and meaning representation. The naturalness is measured by how likely the labeller thinks the utterance is spoken by a real human. In addition to the model output, each labeler is also given the meaning representation and the ground truth. The labelers need to give an integer rating from 1 to 3 (3 being the best) for each criterion. %To reduce bias, we randomly shuffle the order of all models' results for each sample.

Table~\ref{table:human} shows that our AdvNLG model has an apparent lead in both naturalness and accuracy, and the paired t-test shows that the result is statistically significant with p-value smaller than 0.01. And our ablation model -Adv. achieves the lowest score, proving that adversarial training can boost both naturalness and accuracy. 

\section{Conclusion}
\label{conclusion}
In this paper, we propose adversarial training using the Straight-Through Gumbel-Softmax estimator in NLG for task-oriented dialogues. We also propose a two-stage training scheme to further boost the gain in performance. Experimental results show that our model, AdvNLG, consistently outperforms state-of-the-art models in both automatic and human evaluations.

In the future, we plan to apply this method to other conditional generation tasks, e.g. produce a natural utterance containing a given list of keywords.

\section*{Acknowledgement}
We thank the anonymous reviewers for their valuable comments. We thank William Hinthorn for proof-reading the paper.

\bibliography{main}
\bibliographystyle{acl_natbib}
\clearpage

\end{document}